%% file: main.tex
\documentclass{article}

\usepackage[preprint]{corl_2026} 

\usepackage{pifont}

\usepackage{subcaption}
\usepackage{graphicx}
\usepackage{multirow}
\usepackage[ruled,vlined]{algorithm2e}
\usepackage{amsmath}
\usepackage{cleveref}
\usepackage{fontawesome5}
\usepackage{tabularx}
\usepackage{array}
\usepackage{booktabs}
\newcolumntype{C}{>{\centering\arraybackslash}X}
\usepackage{tikz}
\definecolor{mydarkblue}{rgb}{0,0.08,0.45}
\usetikzlibrary{
  backgrounds,
  positioning,
  arrows.meta,
  shadows,
  calc,
  fit,
  shapes.geometric
}

\title{EgoTrack3D: A Modular Framework for Egocentric 3D Object Tracking}

\author{
  \textnormal{Jan Kulik}\thanks{Equal contribution} \\
  ETH Zürich \\
  \texttt{jkulik@ethz.ch} \\
  \and
  Bjarni Dagur Thor Kárason\footnotemark[1] \\
  ETH Zürich \\
  \texttt{bkarason@ethz.ch} \\
  \and
  Yung-Hsu Yang \\
  ETH Zürich \\
  \texttt{yunghsu.yang@inf.ethz.ch} \\
  \and
  Boyang Sun \\
  ETH Zürich \\
  \texttt{boysun@ethz.ch} \\
  \and
  Marc Pollefeys \\
  ETH Zürich \& Microsoft Research\\
  \texttt{marc.pollefeys@inf.ethz.ch} \\
  \and
  Xi Wang \\
  ETH Zürich \& TUM \& MCML \\
  \texttt{xi.wang@inf.ethz.ch} \\
}

\begin{document}
\maketitle


	



	

\input{sec/0_abstract}
\input{sec/1_introduction}
\input{sec/2_related_work}
\input{sec/3_method}
\input{sec/4_experiments}
\input{sec/5_results}
\input{sec/6_limitations}
\input{sec/7_conclusion}


\acknowledgments{This work was supported by the Swiss National Science Foundation Advanced Grant 216260: Beyond Frozen Worlds: Capturing Functional 3D Digital Twins from the Real World.}


\bibliography{main}  

\clearpage
\appendix
\input{sec/8_appendix}

\end{document}

%% file: sec/0_abstract.tex
\begin{abstract}
Understanding 3D scenes from egocentric video is fundamental for robotics and autonomous navigation, yet rapid viewpoint changes and partial occlusions make building structured representations challenging. Existing 3D tracking and scene graph construction methods primarily address explicit interactions or assume static scenes, limiting their ability to capture complex dynamics. We introduce EgoTrack3D, a modular framework that reconstructs and maintains a dynamic 3D scene representation directly from egocentric RGB video. The framework lifts 2D segmentation masks into a global 3D coordinate frame, using a point-based motion scoring mechanism alongside a voxel-based merging heuristic to associate object tracks. EgoTrack3D maintains accurate representations over time, achieving an 11\% improvement in percentage of correct locations (PCL) relative to the strongest baseline on the Aria Digital Twin (ADT) dataset, while addressing the more general setting of persistent 3D tracking for both static and dynamic objects. Furthermore, to demonstrate the system's robustness under degraded conditions that simulate real-world deployment constraints, we replace dense depth maps with sparse 3D bounding box estimation and integrate interaction-guided dynamic association, enabling EgoTrack3D to maintain accurate spatial representations despite noisy observations.
\end{abstract}

\keywords{Egocentric 3D Perception, 3D Object Tracking, Dynamic Scene Reconstruction}

%% file: sec/1_introduction.tex
\section{Introduction}
\label{sec:intro}

Understanding 3D scenes is fundamental for robotics, autonomous navigation, and embodied AI. Robots must build structured representations of their surroundings to plan safe, goal-directed motion. Scene graphs provide such a structure by describing objects, their attributes, and their spatial or semantic relationships, enabling efficient reasoning over complex environments.

Although 2D scene graphs have advanced high-level reasoning in vision tasks such as image retrieval and visual question answering~\citep{johnson2015image, krishna2017visual}, 3D scene graphs extend these ideas to real-world perception and interaction~\citep{hughes2022hydra, gu2024conceptgraphs, gorlo2024long}. However, existing approaches typically assume static scenes~\citep{armeni20193d, kim20193d, wu2021scenegraphfusion, koch2024open3dsg} or rely on precomputed representations to track dynamic objects~\citep{behrens2025lost}.

Egocentric videos provide rich, temporally continuous observations from a first-person perspective, but introduce major challenges, including rapid viewpoint changes, partial occlusions, truncated fields of view, and objects frequently entering or leaving the scene. Recent methods leverage egocentric video for 3D object tracking~\citep{zhao2024instance, Plizzari2025OSNOM, bhalgat20243d}, but they address related yet narrower settings: some focus on predefined or interacted objects, while others refine 2D tracks or localize objects using compact 3D representations. Boxer~\citep{boxer2026} lifts 2D detections into globally consistent 3D boxes, but is designed for static object localization rather than dynamic egocentric tracking. As a result, existing methods do not fully address persistent full-scene tracking of all observed static and dynamic objects.

To address these challenges, we introduce EgoTrack3D, a modular framework for constructing and updating a 3D scene representation directly from egocentric video. EgoTrack3D lifts segmentation masks into a shared 3D coordinate frame using depth maps, camera intrinsics, and poses, and tracks both static and moving objects using geometric and visual features. Object motion is detected through a point-based 3D motion scoring mechanism built on CoTracker3~\citep{karaev2024cotracker3}, so that observations of static objects can be aggregated while moving objects are updated more dynamically. Object tracks are merged using a 3D voxel-based heuristic to prune duplicates. Unlike existing 3D object tracking methods, EgoTrack3D captures an object's entire point cloud, providing richer cues for 3D tracking.

A practical challenge is that dense metric depth is difficult to obtain reliably in unconstrained egocentric videos. While the Aria Digital Twin (ADT) dataset~\citep{pan2023aria} provides ground-truth depth, directly replacing it with monocular predictions leads to unstable object point clouds and unreliable 3D association. We therefore treat EgoTrack3D as a modular framework rather than a fixed pipeline: when dense depth is available, masks are lifted through per-pixel depth; when only sparse geometry is available, the lifting module can be replaced by a learned 3D box lifter.

Our contributions are: \textbf{(i)} a modular pipeline for lifting 2D segmentations into temporally consistent 3D object tracks; \textbf{(ii)} a motion-aware association mechanism combining geometry and appearance; \textbf{(iii)} a sparse-input variant using BoxerNet-based 3D boxes and interaction-guided dynamic association; and \textbf{(iv)} evaluation on ADT and HD-EPIC showing improved 3D tracking consistency under dense and sparse inputs.


%% file: sec/2_related_work.tex
\section{Related work}
\label{sec:related_work}
Recent open-world video models enable object-level perception: the SAM model family~\citep{kirillov2023segment,ravi2024sam,carion2025sam3segmentconcepts} provides promptable segmentation and mask propagation; DEVA, Track Anything, and SAM-PT~\citep{cheng2023tracking,yang2023track,rajivc2025segment} improve temporal consistency. Despite these advances, purely 2D methods still struggle to preserve object identity through long occlusions, large viewpoint changes, and hand-object interactions.

Incorporating 3D information has recently improved egocentric object tracking and localization. IT3DEgo~\citep{zhao2024instance} projects 2D detections into 3D using depth and pose, but remains limited to predefined interactive objects. OSNOM~\citep{Plizzari2025OSNOM} introduced the Lift--Match--Keep strategy to retain out-of-sight objects via 3D lifting, while EgoSeg3D~\citep{bhalgat20243d} refines 2D tracks from pretrained video segmentation models using 3D cues. Boxer~\citep{boxer2026} lifts open-vocabulary 2D detections from posed images into globally consistent static 3D bounding boxes, demonstrating the value of 3D lifting for static object localization. However, these methods either address static 3D localization or track a restricted set of task-relevant objects, rather than maintaining persistent 3D representations for all visible objects in dynamic egocentric observations. EgoTrack3D differs by tracking all visible objects using both geometric and appearance features, constructing full 3D scene representations directly from egocentric RGB video.

Depth estimation provides essential 3D information for 3D lifting, with recent monocular methods predicting metric depth~\citep{bhat2023zoedepth,yin2023metric3d,piccinelli2025unik3d}, relative depth~\citep{yang2024depthv1,yang2024depthv2,ke2024repurposing}, or temporally consistent video depth~\citep{shao2025learning,hu2025depthcrafter,chen2025video}. However, egocentric motion remains challenging, so our dense setting uses ADT ground-truth depth to isolate tracking and scene reconstruction from depth-prediction errors.

Point tracking has progressed from independent point trajectories~\citep{harley2022particle,doersch2022tap} to joint transformer-based tracking with cross-track attention~\citep{karaev2024cotracker}. We use the more efficient CoTracker3~\citep{karaev2024cotracker3} to derive per-point 3D motion scores for selective object updates.

Finally, 3D scene graphs provide structured representations linking geometry, semantics, and relationships. Early works~\citep{kim20193d, armeni20193d, rosinol20023d} modeled static environments, later extended by SceneGraphFusion~\citep{wu2021scenegraphfusion}, ConceptGraphs~\citep{gu2024conceptgraphs}, and Open3DSG~\citep{koch2024open3dsg} to incrementally build or predict 3D graphs from RGB-D data. Lost\&Found~\citep{behrens2025lost} introduced dynamic tracking but relies on a precomputed static map. While EgoTrack3D does not produce a scene graph, it generates a 3D representation of the environment entirely from egocentric video, without requiring any prior environment model. This representation can be turned into a scene graph by incorporating semantics and inferring relationships.

%% file: sec/3_method.tex
\section{Method}
\label{sec:method}

Given an egocentric video with $T$ RGB frames, our goal is to construct a temporally consistent 3D representation of the environment containing both static and dynamic objects observed throughout the sequence. Unlike prior works that either track only a restricted set of interacted or task-relevant objects, localize static objects without maintaining temporal tracks, or represent objects only by their centers, our method estimates and tracks full 3D bounding boxes for all visible objects at every frame $1 \leq t \leq T$.

EgoTrack3D is formulated as a modular tracking framework whose lifting module can be instantiated with different sources of 3D information. We consider two variants. The dense variant uses per-pixel metric depth to lift segmentation masks into object point clouds and is used for controlled evaluation on ADT. The sparse variant replaces dense lifting with BoxerNet-based~\citep{boxer2026} 3D box prediction from sparse geometry, enabling evaluation under more realistic inputs where dense depth is unavailable or unreliable.

\subsection{EgoTrack3D-Dense}

Our pipeline is shown in \Cref{fig:entire_pipeline_diagram}. It consists of
\textbf{(i)} per-frame segmentation and depth estimation,
\textbf{(ii)} motion detection via CoTracker3~\citep{karaev2024cotracker3},
\textbf{(iii)} appearance embedding extraction using MASA~\citep{li2024matching},
\textbf{(iv)} 3D lifting of segmented masks, and
\textbf{(v)} 3D multi-object tracking and merging.

\input{tikz_figures/pipeline}

\subsubsection{Detecting moving objects}
To determine whether an object is moving at frame $t$, we track an $N \times N$ grid of points over $\Delta t$ frames using CoTracker3~\citep{karaev2024cotracker3}. A tracked 3D point $p_{n,t}$ yields a trajectory $\mathbf{P}_{n,t}=\{p_{n,t,i}\}_{i=1}^{\Delta t}$, where $\Delta p_{n,t,i}$ denotes the 3D displacement of point $n$ between consecutive frames. Its motion and direction scores are:

\begin{equation}
    m_{n,t} = \sum_{i = 2}^{\Delta t} || \Delta p_{n,t,i} ||_2,
    \qquad
    \mathbf{d}_{n,t} = \frac{1}{\Delta t-1} \sum_{i=2}^{\Delta t} \frac{\Delta p_{n,t,i}}{|| \Delta p_{n,t,i} ||_2}.
\end{equation}

Per-object scores are obtained by averaging over all points belonging to mask $\mathcal{M}_j$:
\begin{equation}
    m_t^{\mathcal{M}_j} = \frac{1}{|\mathbf{P}(\mathcal{M}_j)|} \sum_{p_{n,t} \in \mathbf{P}(\mathcal{M}_j)} m_{n,t},
    \qquad
    \mathbf{d}_t^{\mathcal{M}_j} = \frac{1}{|\mathbf{P}(\mathcal{M}_j)|} \sum_{p_{n,t} \in \mathbf{P}(\mathcal{M}_j)} \mathbf{d}_{n,t}.
\end{equation}
An object is classified as moving if $m_t^{\mathcal{M}_j} \ge \tau_m$ and $|\mathbf{d}_t^{\mathcal{M}_j}| \ge \tau_d$. We use $N = 60$, $\Delta t = 10$, $\tau_m = \tau_d = 0.1$, and only consider points within 1 m of the camera.

\subsubsection{Instance appearance features}
We extract per-object appearance embeddings using MASA~\citep{li2024matching}.
For each segmentation mask, we use its 2D bounding box to pool features from MASA's backbone and obtain an object embedding $\mathbf{q}$. This embedding similarity provides robust cross-frame association signals.

\subsubsection{3D detection via lifting}
Each 2D image point $\mathbf{u} = [u, v, 1]^\top$ is projected from image space to world coordinates using its depth $d$, the camera intrinsics matrix $\mathbf{K}$, and the camera-to-world pose $\mathbf{T}_{cw}$, using
\begin{equation}
    \begin{bmatrix} X & Y & Z & 1 \end{bmatrix}^\top = \mathbf{T}_{cw} \cdot \begin{bmatrix} d \cdot \mathbf{K}^{-1} \mathbf{u} \\ 1 \end{bmatrix}.
\end{equation} 
For each object, we gather its 3D points, apply voxel downsampling, and fit an oriented bounding box using principal component analysis (PCA) over the object's point cloud $P$, yielding $\text{BOX}_{\text{PCA}}(P)$.

\subsubsection{Tracking objects in 3D}
Let $\mathcal{T}_i^{t-1} = \{P_i^{t-1},o_i^{t-1},\mathbf{q}_i^{t-1}\}$ denote an existing track and $\mathcal{D}_j^t = \{P_j^t,o_j^t,\mathbf{q}_j^t\}$ a new detection. Here, $P$ is the object's 3D point cloud, $o$ is its oriented bounding box, and $\mathbf{q}$ is its visual feature embedding. We define the matching cost
\begin{align}
    \label{eq:matching_cost}
    c_{ij} = \lambda_{\text{iou}} \cdot (1 - \text{IoU}(i, j)) + \lambda_{\text{chamfer}} \cdot \text{Chamfer}(i, j) + \lambda_{\text{feat}} \cdot (1 - s(i, j)),
\end{align}
and solve the assignment using the Hungarian algorithm. The 3D intersection over union (IoU) captures overlap, Chamfer distance incorporates shape similarity and distance, and the feature term ensures visual consistency.

Matched pairs are merged as
\begin{equation}
    P_i^t = \begin{cases} P_i^{t-1} \cup P_j^t & \text{if $j$ is static at time $t$} \\ P_j^t & \text{otherwise} \end{cases},
    \quad
    \mathbf{q}^t = \alpha \cdot \mathbf{q}_i^{t-1} + (1 - \alpha) \cdot \mathbf{q}_j^t,
\end{equation}
and we recompute $o_i^t=\text{BOX}_{\text{PCA}}(P_i^t)$. Unmatched detections initialize new tracks and unmatched tracks are propagated to the next frame.

\subsubsection{Track merging}
To eliminate duplicates arising from partial or occluded views, we merge tracks whose 3D point clouds overlap substantially. Let $V(P)$ denote the occupied 3D voxels of a point cloud. Tracks $i,j$ are merged if
\begin{equation}
    \frac{|V(P_i^t)| + |V(P_j^t)| - |V(P_i^t \cup P_j^t)|}{|V(P_i^t)| + |V(P_j^t)|} > \tau,
\end{equation}
indicating significant overlap between the point clouds.

\subsection{EgoTrack3D-Sparse}

The sparse variant replaces dense mask lifting with BoxerNet-based 3D box prediction. For each frame, we obtain instance masks with CropFormer~\citep{qi2023highqualityentitysegmentation} and convert them to 2D box prompts. Given the RGB image, camera calibration, pose, sparse world points, and prompts, BoxerNet predicts one oriented 3D bounding box per object. We retain high-confidence predictions and discard prompts near image borders or overlapping excluded hand regions, which often produce unstable boxes in egocentric video.

Since sparse-depth box predictions are less naturally tied to per-object point clouds, we use hand-object interaction as the main dynamic-object cue. We run Hands23~\citep{hands23} on each frame to detect hands and their interacted objects. A segmentation mask is marked as dynamic when it overlaps a Hands23 primary-object mask above a threshold. This design targets a common and important source of object motion in egocentric scenes: objects manipulated by the camera wearer.

Because BoxerNet is designed for lifting 2D detections into static 3D boxes, its predictions for manipulated or hand-held objects can be noisy and temporally unstable. To address this, we introduce a dynamic mask associator. When a mask is classified as dynamic, we initialize a short-term SAM2~\citep{ravi2024sam} track from the corresponding mask and propagate it for $K$ frames. For each subsequent frame, detections whose masks overlap the propagated SAM2 mask above a threshold are hard-associated to the same object identity before geometric assignment is performed. This creates a temporary 2D anchor for manipulated objects whose 3D box estimates are too noisy for reliable box-level matching.

The tracker assigns BoxerNet detections to existing tracks using the cost in Eq.~\eqref{eq:matching_cost}. Static tracks are smoothed by confidence-weighted box interpolation, while dynamic tracks use the current prediction to avoid fusing objects across different physical locations. Duplicate tracks are merged using 3D box and appearance similarity, except when protected by distinct dynamic-mask identities.

%% file: tikz_figures/pipeline.tex
\begin{figure}[htbp]
    \centering
    \resizebox{\textwidth}{!}{%
    \begin{tikzpicture}[
        node distance=.25cm and .40cm,
        box/.style={
            draw, 
            thick, 
            rounded corners, 
            align=center, 
            minimum width=1.75cm, 
            minimum height=1cm,
            text width=1.75cm,
            font=\scriptsize,
            fill=white,
            drop shadow
        },
        ->, >=Stealth
    ]

    \node[box] (segmentation) {
        \includegraphics[width=1.75cm]{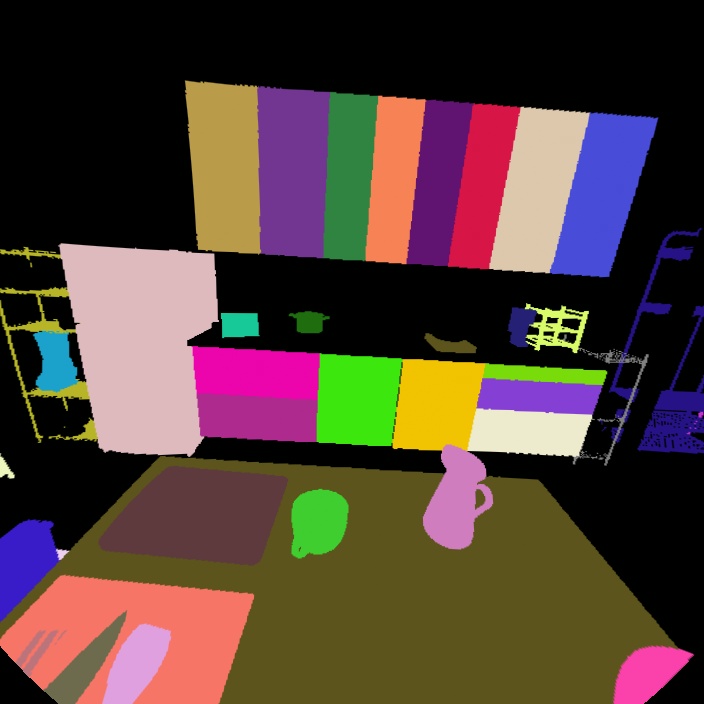} 
        Generate segmentation masks
    };
    \node[box, below=of segmentation] (depth) {
        \includegraphics[width=1.75cm]{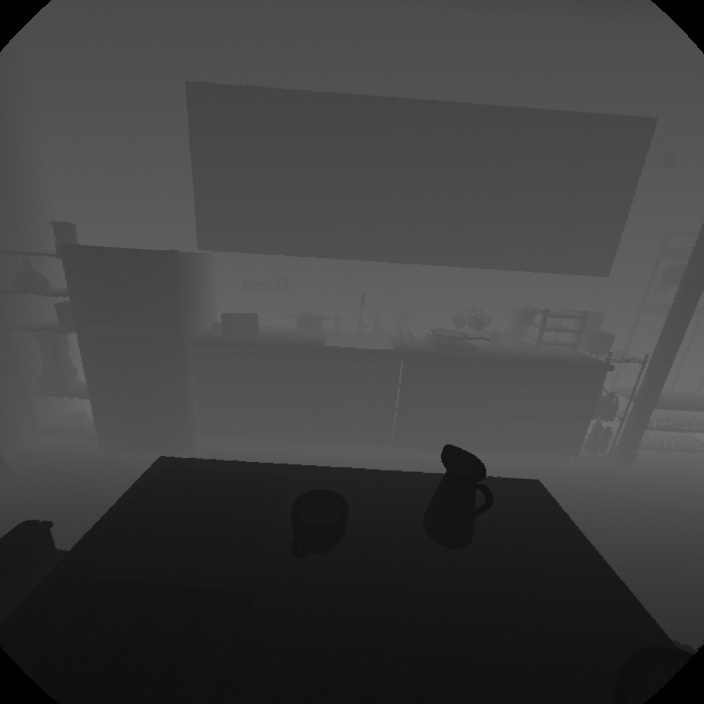} \\[0.5em]
        Generate depth maps
    };
    \node[box, right=of segmentation] (cotracker) {
        \includegraphics[width=1.75cm]{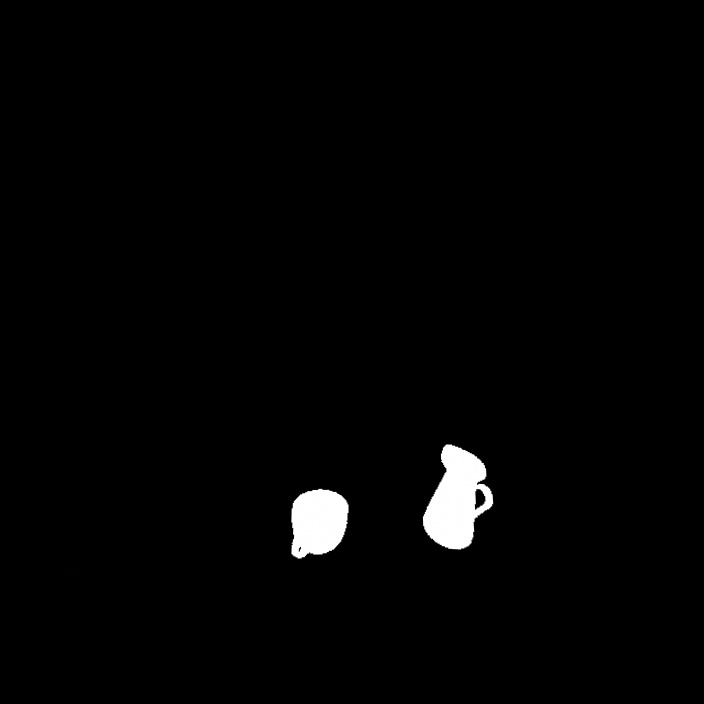} \\[0.5em]
        Find moving objects using CoTracker3
    };
    \node[box, right=of depth] (masa) {
        Generate MASA features
    };

    \draw (segmentation) -- (cotracker);
    \draw (segmentation) -- (masa);
    \draw[->, bend right=15] (depth) to (cotracker);

    \begin{scope}[on background layer]
        \node[draw=gray, thick, dashed, rounded corners, inner sep=12pt, fill=gray!10,
              fit=(segmentation)(depth)(masa)(cotracker)] (preprocessing_group) {};
    \end{scope}
    \node[anchor=south west, font=\scriptsize\itshape\color{gray}] 
          at (preprocessing_group.south west) {\faClock for every frame $t$};
    \node[anchor=south, font=\bfseries\small\color{gray}] 
          at (preprocessing_group.north) {Preprocessing};

    \node[box, left=.5cm of preprocessing_group.west] (rgb) {
        \includegraphics[width=1.75cm]{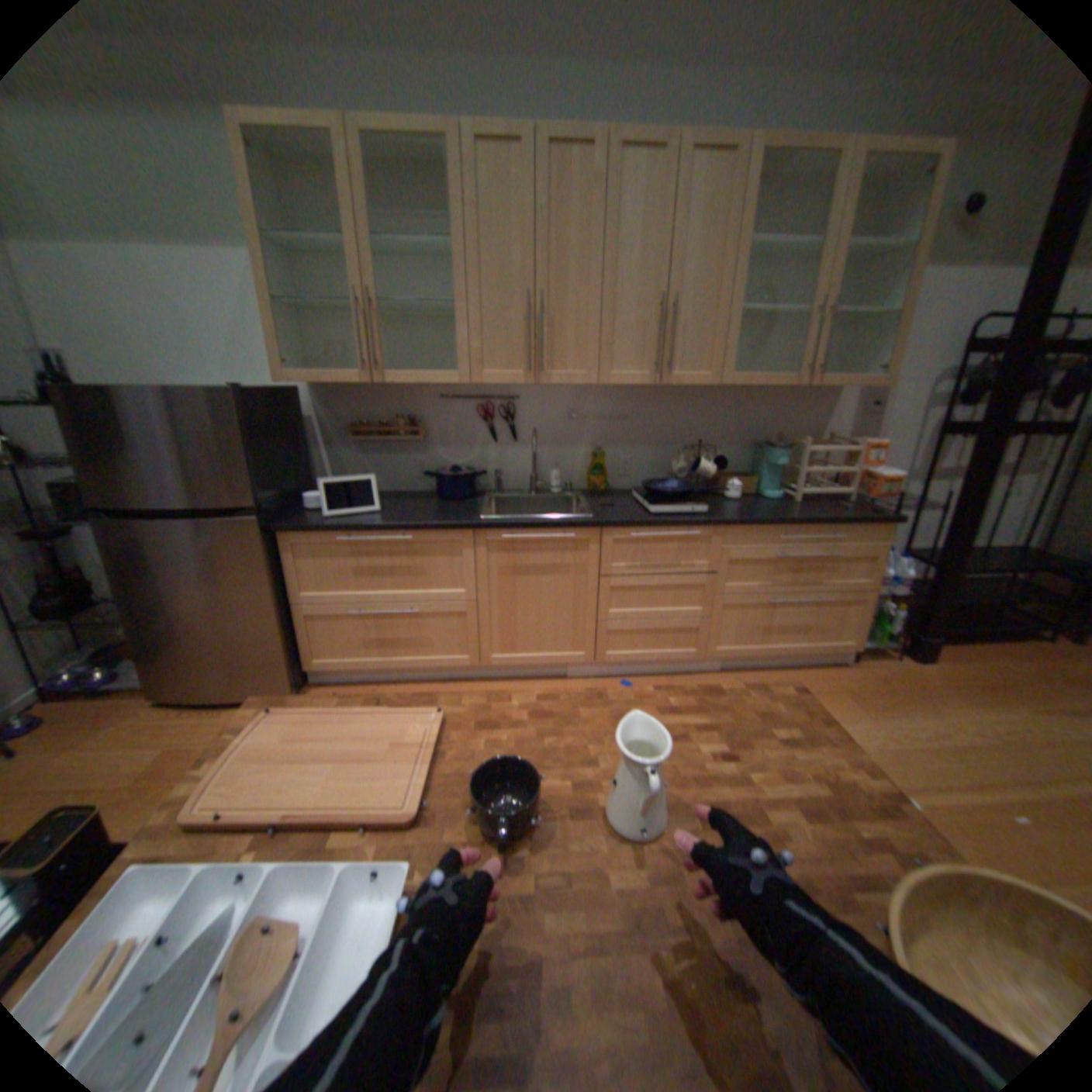} \\[0.5em]
        RGB Aria data
    };
    \node[font=\bfseries\small\color{gray}] at ([yshift=0.5cm]rgb.north) {Input};
    \draw (rgb) -- (preprocessing_group);

    \node[box, anchor=north west] (filter_segmentation) 
         at ($(cotracker.north east)+(1.25cm,0)$) {Filter segmentation masks};

    \node[box, below=of filter_segmentation] (lift_boxes) {
        \includegraphics[width=1.75cm]{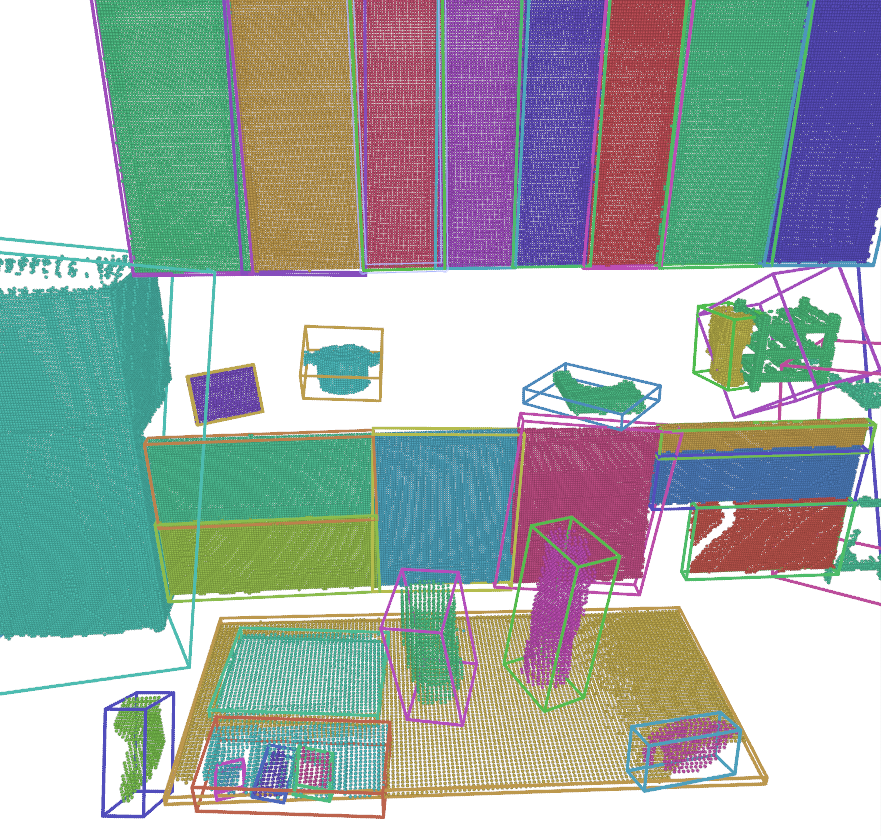} \\[0.5em]
         Lift masks to 3D and extract bounding boxes
    };

    \node[box, below=of lift_boxes] (match) {Match and combine};

    \node[box, right=of match] (tracks) {
        Existing object tracks from $t - 1$
    };
    \node[box, above=of tracks] (merge_tracks) {
        Merge tracks
    };

    \draw (filter_segmentation) -- (lift_boxes);
    \draw (lift_boxes) -- (match);
    \draw (tracks) -- (match);
    \draw (match) -- (merge_tracks);

    \begin{scope}[on background layer]
        \node[draw=gray, thick, dashed, rounded corners, inner sep=12pt, fill=gray!10,
              fit=(filter_segmentation)(lift_boxes)(match)(tracks)(merge_tracks)] (tracking_group) {};
    \end{scope}
    \node[anchor=south west, font=\scriptsize\itshape\color{gray}] 
          at (tracking_group.south west) {\faClock for every frame $t$};
    \node[anchor=south, font=\bfseries\small\color{gray}] 
          at (tracking_group.north) {Tracking};
          
    \draw[->, bend left=15] (segmentation.north) to (filter_segmentation.north);
    \draw (cotracker) -- (match.north west);
    \draw (masa) -- (match.west);
    \draw[->, bend left=7] (depth) to (lift_boxes);

    \node[box, right=.75cm of merge_tracks] (output) {
        All object tracks at time $t$
    };
    \node[anchor=south, font=\bfseries\small\color{gray}] 
          at (output.north) {Output};
    \draw (merge_tracks) -- (output);
    \draw[dashed] (output) -- (tracks);
        
    \end{tikzpicture}
    }
    \caption{\textbf{Pipeline overview for generating 3D dynamic scene representations from egocentric RGB video.} For each frame, the RGB input is processed to generate segmentation masks and depth maps, which are used to extract visual features (MASA) and detect moving objects (CoTracker3). The segmentation masks are then filtered, lifted to 3D bounding boxes, and matched to existing object tracks to maintain consistent identities across the entire video sequence.}
    \label{fig:entire_pipeline_diagram}
\end{figure}

%% file: sec/4_experiments.tex
\section{Experiments}
\label{sec:experiments}

\subsection{Datasets and input settings}
\label{sec:data}

We use the ADT dataset~\citep{pan2023aria} for quantitative evaluation on eight sequences. ADT provides ground-truth camera poses, segmentation masks, depth maps, and oriented 3D bounding boxes. All experiments use camera poses provided by Project Aria, and the image data are devignetted and converted from the Fisheye624 lens model to a standard pinhole camera model.

We also use HD-EPIC~\citep{perrett2025hdepic} for qualitative evaluation on unconstrained real-world egocentric videos. Since HD-EPIC lacks dense metric depth, ground-truth instance masks, and temporally consistent 3D object tracks, we do not report quantitative PCL scores on this dataset.

EgoTrack3D-Dense is evaluated on ADT with ground-truth masks, depth maps, and poses to isolate tracking from perception noise. EgoTrack3D-Sparse removes the dense-depth assumption and uses model-generated masks, sparse point clouds, and poses; it is evaluated quantitatively on ADT and qualitatively on HD-EPIC.

\subsection{Metrics}
\label{sec:metrics}
To evaluate scenes reconstructed using ground-truth depth and masks, we use the percentage of correct locations (PCL) metric introduced by OSNOM~\citep{Plizzari2025OSNOM}, as traditional tracking metrics do not evaluate object tracks that are out of sight. For a distance threshold $R$, a predicted object is correct if
\textbf{(i)} its ID matches the ground-truth object ID, and
\textbf{(ii)} the Euclidean distance between their bounding box centers is $\le R$ meters.

The score $\text{PCL}_R$ is the ratio between the number of correct predictions and the total number of unique ground-truth and predicted IDs.
This penalizes both ID switches and missed or duplicate detections. For instance, if a ground-truth object is lost and later reappears under a new predicted ID, both predictions are considered incorrect.

When reconstructing dynamic scenes using predicted segmentation masks, the correspondence between predicted and ground-truth objects is unknown. We therefore evaluate overall 3D scene reconstruction quality using both the F1 score and Average Precision (AP). We treat all detections as valid and assume them to be above a fixed confidence threshold. In this setting, F1 provides a direct measure of scene completeness and correctness.

At each timestep $t$, only ground-truth objects visible up to $t$ are considered, ensuring metric comparability between sequences with different visibility patterns.

\subsection{3D tracking baselines}
\label{sec:3d_tracking_baselines}

We compare against IT3DEgo~\citep{zhao2024instance}, OSNOM~\citep{Plizzari2025OSNOM}, EgoSeg3D~\citep{bhalgat20243d}, and Boxer~\citep{boxer2026}, adapting each method to ADT using ground-truth poses, intrinsics, segmentation masks, and depth where applicable. IT3DEgo is evaluated in its Single-View Object Enrollment (SVOE) setting and is given one enrollment box per object. OSNOM associates lifted detections using 3D location and appearance features. EgoSeg3D is evaluated with and without ground-truth object-ID cues. Boxer is evaluated using the post-processed scene-level object set returned by the released pipeline in the online tracking setting. Additional adaptation details are provided in Appendix~\ref{app:baseline_adaptations}.

%% file: sec/5_results.tex
\section{Results}
\label{sec:results}

\subsection{Dense-input tracking on ADT}
\label{sec:dense_input_tracking}

We evaluate all methods using ground-truth segmentation masks and depth maps to isolate tracking performance from segmentation noise. For the CoTracker3 motion module, we select a lookahead window of $W=10$ based on ADT ground-truth trajectories, obtaining $48.96\%$ precision and $94.69\%$ recall for detecting objects whose 3D box centers move by at least $0.10\,m$ over the next $\Delta t=10$ frames. \Cref{tab:average_pcl_scores} reports the mean PCL scores at the final timestep across all sequences. As expected, EgoSeg3D with $\alpha_c = 10^4$ outperforms both EgoSeg3D ($\alpha_c = 0$) and OSNOM due to access to ground-truth object IDs during association. When $\alpha_c = 0$, EgoSeg3D behaves similarly to OSNOM but inherits DEVA’s unreliable track IDs in cluttered egocentric scenes, leading to reduced consistency. For Boxer, we report final-scene PCL using the post-processed output returned by the released pipeline, hence we omit it from temporal tracking curves. Moreover, Boxer is designed for static object fusion rather than dynamic-object tracking.

\begin{table}[ht]
    \centering
    \caption{\textbf{Average PCL scores (\%) for each method over all videos using ground-truth input data.} ($^\dagger$ IT3DEgo knows how many objects to track and therefore never generates duplicate object tracks. $^\ddagger$ Boxer tracks static scene objects and does not maintain persistent dynamic-object identities.)}

    \resizebox{\textwidth}{!}{%
    \begin{tabular}{@{}lcccccc|c@{}}
        \toprule
        \textbf{Method} & $\textbf{PCL}_{0.1}$ & $\textbf{PCL}_{0.2}$ & $\textbf{PCL}_{0.3}$ & $\textbf{PCL}_{0.6}$ & $\textbf{PCL}_{0.9}$ & $\textbf{PCL}_{1.2}$ & \textbf{Average} \\
        \midrule
            IT3DEgo$^\dagger$            & 23.94 & 30.07 & 33.88 & 40.15 & 44.51 & 46.72 & 36.55 \\
            \midrule
            Boxer$^\ddagger$             & 40.89 & 55.72 & 58.70 & 60.98 & 61.33 & 61.56 & 56.53 \\
            \midrule
            OSNOM                        &  8.34 & 13.11 & 16.54 & 23.93 & 30.49 & 36.05 & 21.41 \\
            EgoSeg3D ($\alpha_c = 0$)    &  4.51 &  7.03 &  8.72 & 12.20 & 14.64 & 15.51 & 10.44 \\
            EgoSeg3D ($\alpha_c = 10^4$) & 13.23 & 19.19 & 21.79 & 26.54 & 28.93 & 30.28 & 23.33 \\
            EgoTrack3D-Dense (no merge)  & 30.42 & 32.91 & 34.25 & 35.96 & 36.21 & 36.38 & 34.36 \\
            EgoTrack3D-Dense (merge)     & \textbf{56.28} & \textbf{60.14} & \textbf{62.56} & \textbf{65.88} & \textbf{66.48} & \textbf{66.72} & \textbf{63.01} \\
        \bottomrule
    \end{tabular}
    }
    \label{tab:average_pcl_scores}
\end{table}

Our method, EgoTrack3D-Dense, achieves the best overall performance. Its track-merging heuristic primarily reduces duplicate tracks rather than directly improving localization. Although occasional overmerges occur when \mbox{(near-)}overlapping point clouds are incorrectly matched, the net effect is strongly positive. The heuristic improves average PCL by 83.38\%.

\begin{figure}[h]
    \centering
    \includegraphics[width=0.92\textwidth]{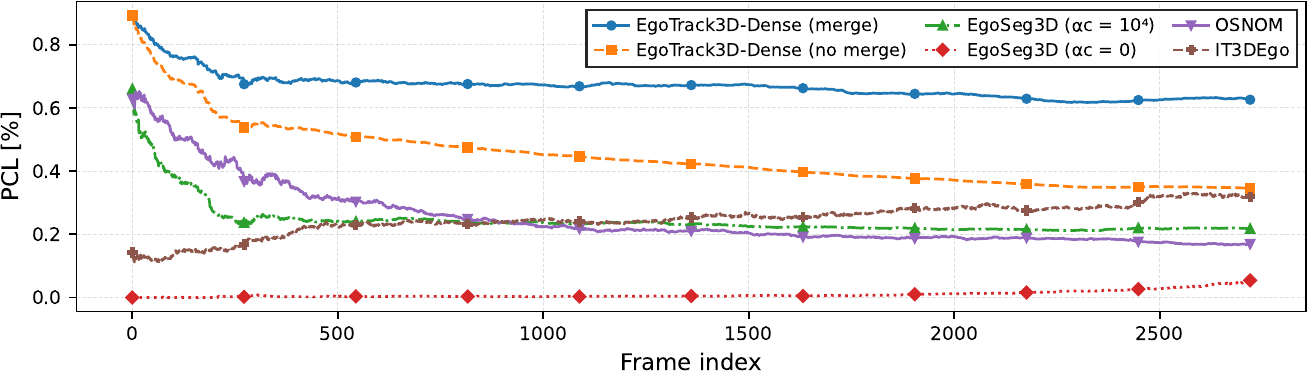}
    \caption{\textbf{$\text{PCL}_{0.3}$ over time for EgoTrack3D-Dense and baselines on ADT.} Boxer is omitted because its output is a post-processed scene-level object set rather than per-frame dynamic tracks.}
    \label{fig:pcl_by_method}
\end{figure}

\Cref{fig:pcl_by_method} shows the evolution of $\text{PCL}_{0.3}$ over time. PCL generally decreases as errors and duplicate tracks accumulate, except for IT3DEgo, which delays initialization until the object is clearly visible, and EgoSeg3D ($\alpha_c = 0$), which struggles with consistent localization but recovers slightly toward the end. Disabling EgoTrack3D's merging heuristic causes a sharper decay, confirming its role in maintaining temporal consistency.

\subsection{Sparse-input adaptation on ADT}
\label{sec:sparse_input_adaptation}

We evaluate EgoTrack3D-Sparse in a setting where dense metric depth and ground-truth masks are unavailable. Instead, the method receives model-generated 2D masks and sparse point clouds, and lifts detections into 3D using BoxerNet. This setting is substantially more challenging than the dense-input setting, since both the 2D object hypotheses and the 3D lifting step are noisy. We compare against Boxer, the closest baseline under sparse-geometry inputs. \Cref{tab:sparse_input_results} reports the average AP and F1 scores across a range of 3D IoU thresholds (0.05--0.50 in increments of 0.05). Following Cube-RCNN/Omni3D~\citep{brazil2023omni3d}, we average AP and F1 over relaxed 3D IoU thresholds, since partial observations often yield under-approximated boxes despite plausible 3D localization.

\begin{table}[ht]
    \centering
    \caption{\textbf{Sparse-input reconstruction and ablations on ADT.} Scores are averaged over 3D IoU thresholds from 0.05 to 0.50 (step size 0.05).}
    \begin{tabular}{lcc}
        \toprule
        \textbf{Method} & $\textbf{AP}@[.05:.05:.50]$ (\%) & $\textbf{F1}@[.05:.05:.50]$ (\%) \\
        \midrule
        Boxer & 14.12 & 29.88 \\
        \midrule
        EgoTrack3D-Sparse (full) & \textbf{24.49} & \textbf{40.79} \\
        \quad w/o 2D dynamic anchoring & 24.11 & 39.28 \\
        \quad w/o interaction-guided dynamics & 23.13 & 40.43 \\
        \bottomrule
    \end{tabular}
    \label{tab:sparse_input_results}
\end{table}

EgoTrack3D-Sparse improves F1 from 29.88 to 40.79, showing that temporal tracking remains useful when dense depth is replaced by sparse, noisy 3D observations. The full variant performs best overall, but removing either 2D dynamic anchoring or interaction-guided dynamics has only a modest effect on aggregate metrics. This is expected, since scene-level AP and F1 are dominated by static objects, whereas these components target a narrower but important failure mode: manipulated objects whose noisy 3D box estimates make geometric association unreliable. We therefore interpret these mechanisms primarily as robustness components for real-world egocentric interactions, which we examine qualitatively in \cref{sec:qualitative_real_world}.

\subsection{Qualitative transfer to real-world video}
\label{sec:qualitative_real_world}

We qualitatively evaluate EgoTrack3D-Sparse on HD-EPIC using CropFormer-generated masks, sparse geometry, and BoxerNet-based 3D lifting to assess whether the sparse-input variant produces plausible temporally consistent 3D tracks under realistic perception noise.

\begin{figure}[h]
    \centering
    
    \begin{subfigure}[b]{0.325\textwidth}
        \centering
        \includegraphics[width=\textwidth, height=\textwidth]{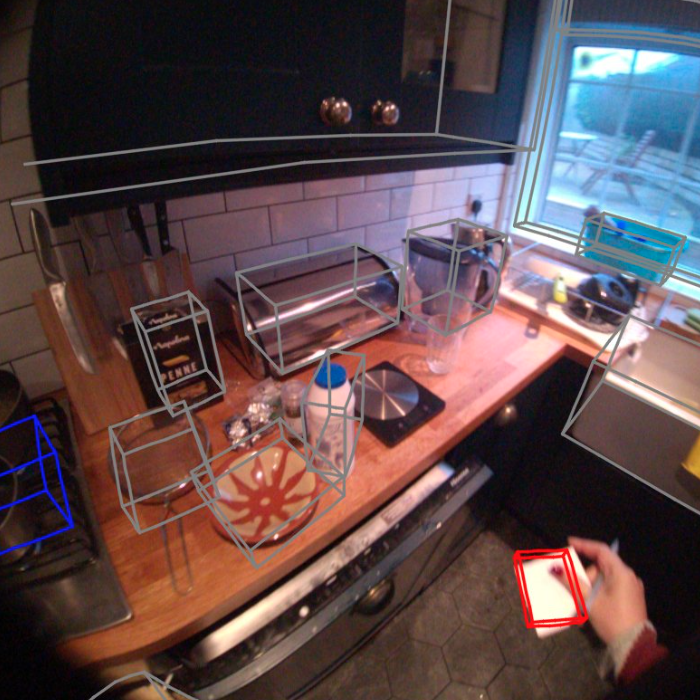}
        \caption{$t=3.9\,s$}
        \label{fig:img1}
    \end{subfigure}
    \hfill
    \begin{subfigure}[b]{0.325\textwidth}
        \centering
        \includegraphics[width=\textwidth, height=\textwidth]{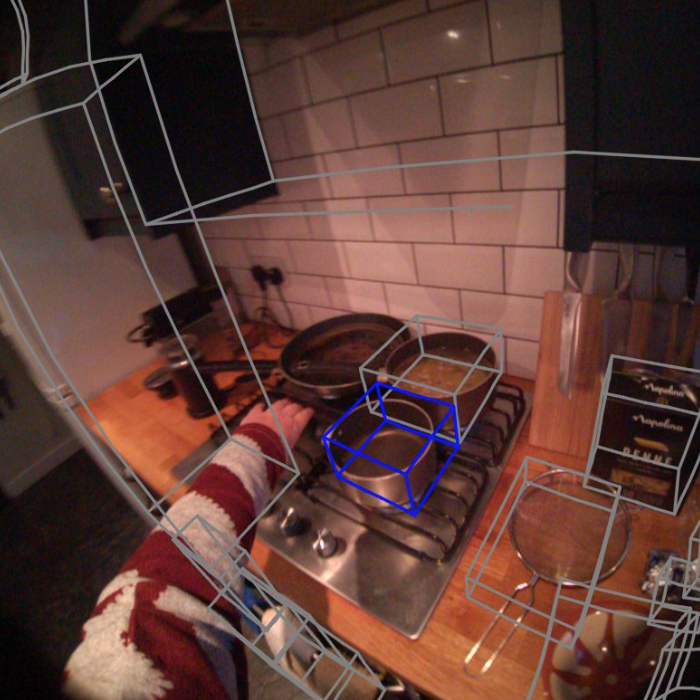}
        \caption{$t=12.3\,s$}
        \label{fig:img2}
    \end{subfigure}
    \hfill
    \begin{subfigure}[b]{0.325\textwidth}
        \centering
        \includegraphics[width=\textwidth, height=\textwidth]{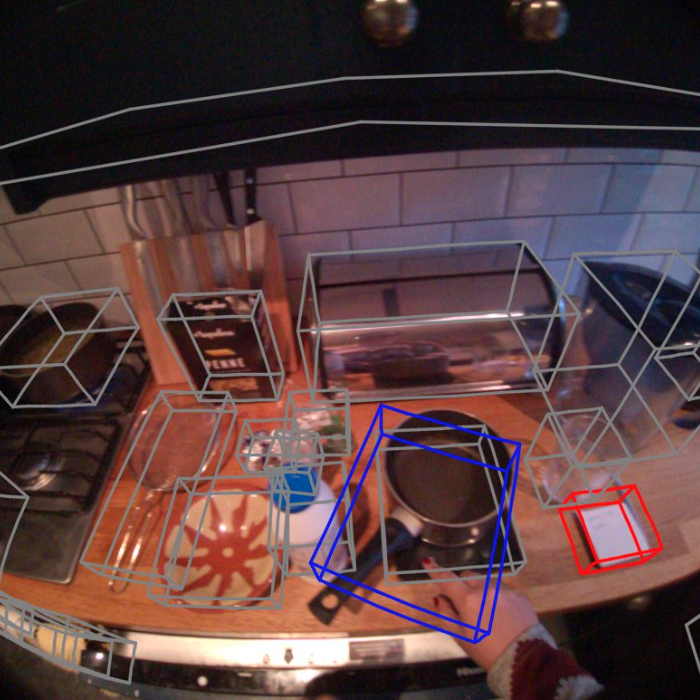}
        \caption{$t=14.5\,s$}
        \label{fig:img3}
    \end{subfigure}
    
    \caption{\textbf{Qualitative real-world tracking with EgoTrack3D-Sparse on HD-EPIC.} Gray boxes show other reconstructed objects, while colored boxes highlight hand-interacted objects. EgoTrack3D-Sparse preserves the notepad (red) and pot (blue) tracks across large viewpoint changes, whereas removing interaction-guided dynamic association causes these objects to be lost or incorrectly associated.}
    \label{fig:hdepic_sparse_qualitative}
\end{figure}

\Cref{fig:hdepic_sparse_qualitative} shows that EgoTrack3D-Sparse reconstructs static scene elements while preserving the identities of manipulated objects across large viewpoint changes. The notepad and pot are challenging because hand-held objects produce noisy BoxerNet predictions; without interaction-guided dynamic association, both objects are incorrectly associated.

%% file: sec/6_limitations.tex
\section{Limitations}
\label{sec:limitations}

EgoTrack3D is a modular pipeline and therefore inherits errors from its perception modules. In the dense setting, it depends on accurate masks, poses, and metric depth. While ADT provides ground-truth depth, replacing it with estimated dense depth remains challenging, as egocentric motion can produce noisy depth, unstable object point clouds, and unreliable 3D associations. Thus, dense results should be interpreted as tracking performance under accurate geometric inputs rather than fully end-to-end real-world perception. The sparse-input variant reduces the dependence on dense depth, but remains sensitive to BoxerNet errors for hand-held or moving objects, whose 3D boxes can be unstable.

Several limitations remain in the tracking itself. Association relies strongly on geometric costs, so re-identification becomes difficult when an object's estimated 3D location changes substantially and visual evidence is insufficient to recover the match. Motion handling can also remove useful history: when an object is classified as moving, prior observations are discarded to avoid outdated geometry, but this can underrepresent large static objects if they are misclassified as dynamic. Future work should explore confidence-aware track management, stronger long-term re-identification via temporally aware or interaction-based association models, and motion-aware 3D box estimation.


%% file: sec/7_conclusion.tex
\section{Conclusion}
\label{sec:conclusion}

We presented EgoTrack3D, a modular framework for constructing temporally consistent 3D object-centric scene representations from egocentric video. EgoTrack3D-Dense uses accurate masks, depth, and poses to lift objects into a shared 3D frame, combining motion scoring, geometric and appearance-based association, and voxel-based track merging to improve tracking consistency on ADT. EgoTrack3D-Sparse replaces dense mask lifting with sparse 3D box estimation from model-generated masks and point clouds, and uses interaction-guided dynamic association to preserve manipulated objects under noisy sparse inputs.

Together, these variants show that EgoTrack3D can operate across both accurate geometric inputs and noisier sparse inputs, while maintaining temporally consistent 3D object tracks. EgoTrack3D moves toward generalizable dynamic 3D scene reconstruction for robotics, augmented reality, and embodied spatial reasoning.

%% file: sec/8_appendix.tex
\section{Baseline adaptation details}
\label{app:baseline_adaptations}

\paragraph{IT3DEgo.}
IT3DEgo~\citep{zhao2024instance} requires one 2D bounding box per object in its Single-View Object Enrollment setting. For each ground-truth object, we automatically select a frame in which the object is clearly visible by choosing the mask with the highest segmentation area-to-perimeter ratio, and use the corresponding 2D bounding box for enrollment. This gives IT3DEgo the number of objects and their video-specific features, meaning that it never produces more tracks than necessary. Since the authors did not release their 3D-guided Kalman filter implementation, we evaluate IT3DEgo using only 2D DINOv2 features for association and lift the resulting tracks to 3D for PCL evaluation.

\paragraph{OSNOM.}
OSNOM~\citep{Plizzari2025OSNOM} receives ground-truth segmentation masks and depth maps. For each detection, it either associates the instance with an existing track or initializes a new one based on the 3D location of the 2D bounding box center and appearance features. This design is conceptually closest to EgoTrack3D-Dense, but OSNOM maintains compact object locations rather than accumulating full object point clouds.

\paragraph{EgoSeg3D.}
EgoSeg3D~\citep{bhalgat20243d} refines 2D tracker outputs using predicted class labels, track IDs, 3D center points, and appearance features. To adapt it to ADT, we modify DEVA to use ground-truth segmentation masks and replace class labels with unique object instance IDs. We evaluate EgoSeg3D with $\alpha_c = 0$, which disables the object-ID cue, and with $\alpha_c = 10^4$, the value recommended by~\citet{bhalgat20243d}, to isolate the effect of
object-ID information.

\paragraph{Boxer.}
Boxer~\citep{boxer2026} lifts 2D bounding boxes from posed images into globally consistent 3D boxes. The released implementation provides an offline fusion mode and an online tracking mode; however, both modes export a post-processed scene-level object set. We therefore evaluate the native exported output rather than extracting intermediate track states. For the dense-input ADT comparison, we replace sparse 3D points with dense ground-truth depth when lifting detections. Since Boxer targets fused static scene objects rather than dynamic-object identities, we include it only in final-scene metrics and omit it from temporal tracking curves.

\section{Additional dense-input results}
\label{app:additional_dense_results}

\begin{figure}[h]
    \centering
    \includegraphics[width=0.92\textwidth]{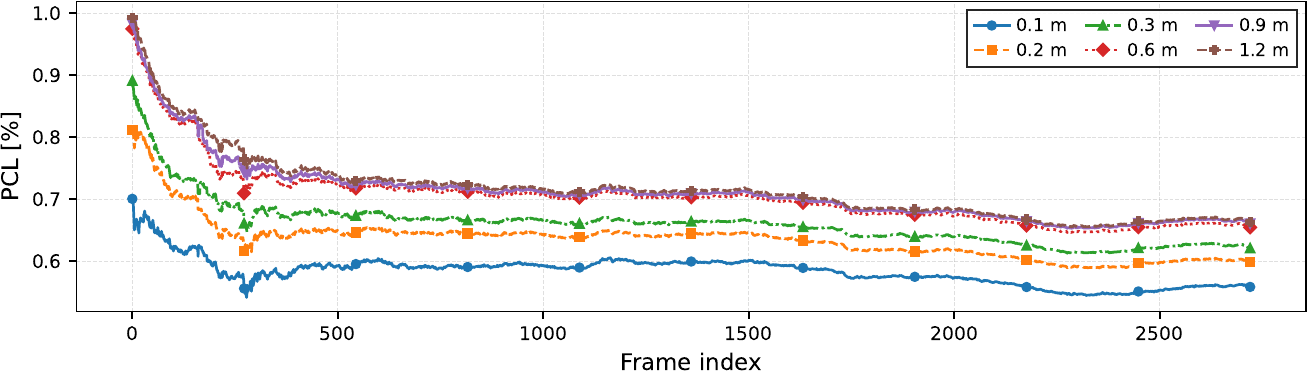}
    \caption{PCL of EgoTrack3D-Dense with the track merging heuristic split by threshold.}
    \label{fig:pcl_by_threshold}
\end{figure}

Figure~\ref{fig:pcl_by_threshold} shows the performance split by PCL radius thresholds. As expected, larger radii lead to higher scores, with diminishing returns beyond $R = 0.6$ m.

Additionally, we present qualitative results on the ADT Meal 132 sequence, comparing reconstructed 3D scene across different settings. Specifically, we visualize the final scene produced by EgoTrack3D-Dense with and without the track merging heuristic, and compare them to the ground-truth scene. We also include the final predicted object locations produced by the 3D tracking baselines. All visualizations are shown from a bird's-eye view.

\begin{figure}[h]
    \centering
    \begin{subfigure}[t]{0.32\textwidth}
        \centering
        \includegraphics[height=5.8cm]{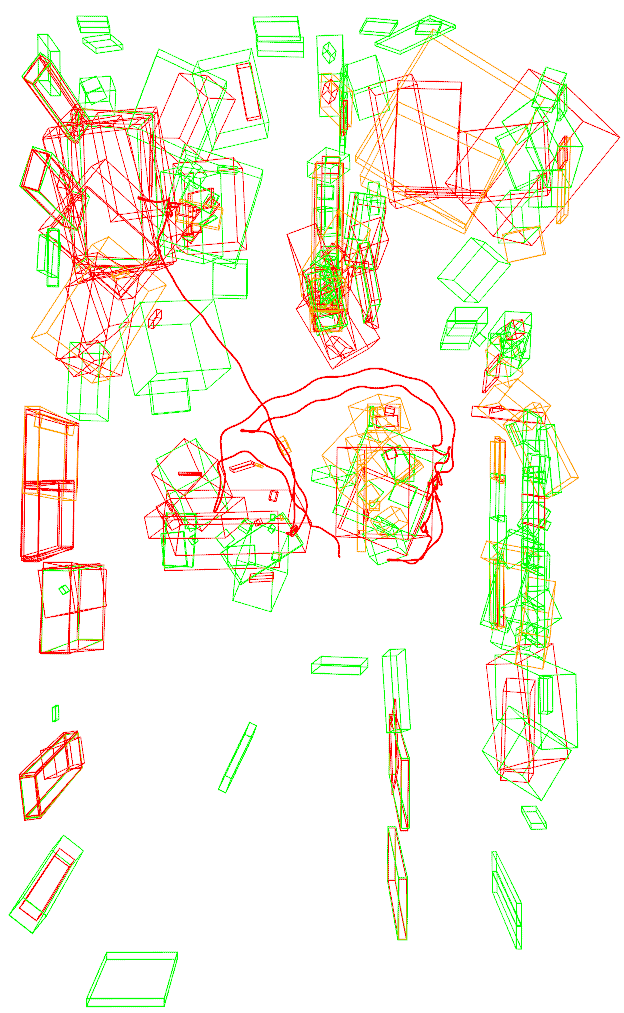}
        \caption{EgoTrack3D without track merging.}
        \label{fig:meal_132_scene_not_merged}
    \end{subfigure}
    \hfill
    \begin{subfigure}[t]{0.32\textwidth}
        \centering
        \includegraphics[height=5.8cm]{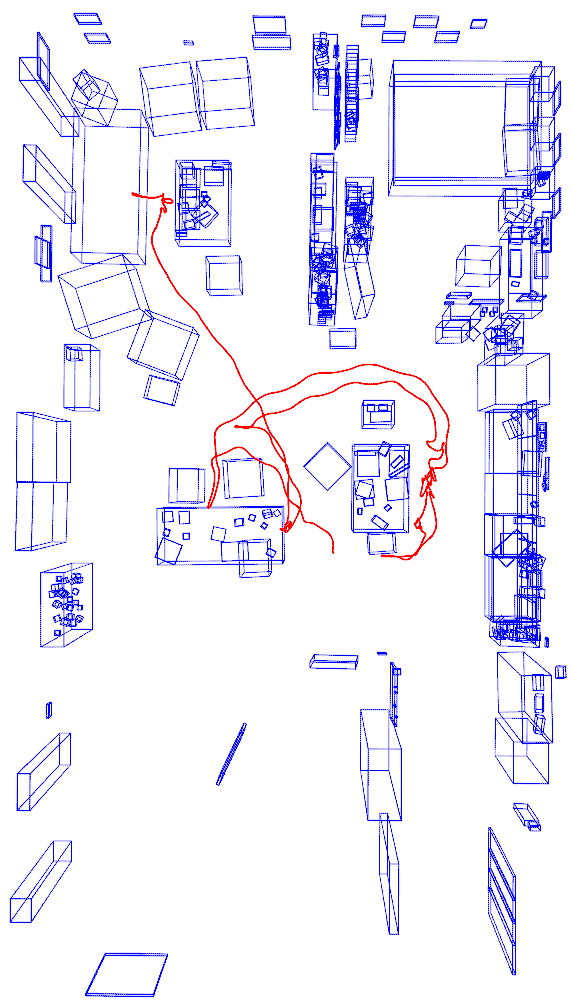}
        \caption{Ground truth.}
        \label{fig:meal_132_gt_scene}
    \end{subfigure}
    \hfill
    \begin{subfigure}[t]{0.32\textwidth}
        \centering
        \includegraphics[height=5.8cm]{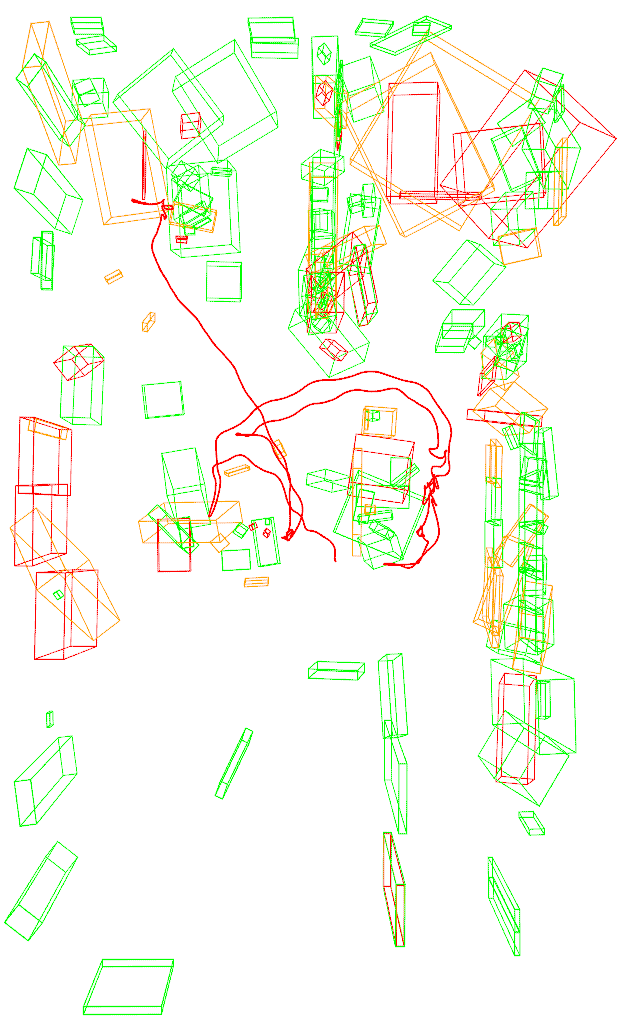}
        \caption{EgoTrack3D with track merging.}
        \label{fig:meal_132_scene_merged}
    \end{subfigure}
    \caption{\textbf{A qualitative comparison of the generated scene representations of the ADT Meal 132 sequence compared to the ground-truth representation.} Green boxes are those that are at most 30 cm away from their ground-truth locations and count as positive detections toward the $\text{PCL}_{0.3}$ metric, while the orange boxes are further away. The red boxes correspond to false-positive duplicate predictions. The agent's trajectory is shown in red.}
    \label{fig:meal_132_scene_comparison}
\end{figure}

\Cref{fig:meal_132_scene_comparison} compares the reconstructed 3D scenes of EgoTrack3D-Dense to the ground truth. Ground-truth bounding boxes are shown in blue, and the agent trajectory in red. In the predicted scenes, green boxes indicate correctly localized objects according to the $\text{PCL}_{0.3}$ metric, orange boxes represent objects displaced by more than 30 cm, and red boxes denote duplicated object tracks.

The ground-truth scene contains 211 visible objects. Without the track merging heuristic, EgoTrack3D-Dense predicts 267 objects, including 96 duplicates. Enabling the heuristic reduces this to 182 objects, with only 28 duplicates. As illustrated in \Cref{fig:meal_132_scene_comparison}, the track merging heuristic effectively eliminates redundant tracks, producing a more compact and accurate representation. However, in rare cases, erroneous associations between nearby but distinct objects cause the heuristic to merge unrelated tracks.